\documentclass[10pt,letterpaper]{article}
\usepackage[top=0.85in,left=2.75in,footskip=0.75in]{geometry}
\usepackage{amsmath,amssymb}
\usepackage{changepage}
\usepackage{textcomp,marvosym}
\usepackage{cite}
\usepackage{nameref,hyperref}
\usepackage[right]{lineno}
\usepackage[nopatch=eqnum]{microtype}
\DisableLigatures[f]{encoding = *, family = * }
\usepackage[table]{xcolor}

\usepackage{array}
\newcolumntype{+}{!{\vrule width 2pt}}
\newlength\savedwidth

\raggedright
\usepackage[aboveskip=1pt,labelfont=bf,labelsep=period,justification=raggedright,singlelinecheck=off]{caption}

\makeatletter
\renewcommand{\@biblabel}[1]{\quad#1.}
\makeatother
\usepackage{lastpage,fancyhdr,graphicx}
\usepackage{epstopdf}
\fancyheadoffset[L]{2.25in}
\fancyfootoffset[L]{2.25in}
\usepackage{subcaption, float, authblk, hyperref}
\begin{document}

\vspace*{0.2in}

\begin{flushleft}

{\Large
 \textbf\newline{Automated Segmentation of Coronal Brain Tissue Slabs for 3D Neuropathology}
}
\newline
\\
Jonathan Williams Ramirez\textsuperscript{1},
Dina Zemlyanker\textsuperscript{1},
Lucas Deden-Binder\textsuperscript{1},
Rogeny Herisse\textsuperscript{1},
Erendira Garcia Pallares\textsuperscript{1},
Karthik Gopinath\textsuperscript{1},
Harshvardhan Gazula\textsuperscript{1},
Christopher Mount\textsuperscript{2},
Liana N. Kozanno\textsuperscript{2},
Michael S. Marshall\textsuperscript{2},
Theresa R. Connors\textsuperscript{2},
Matthew P. Frosch\textsuperscript{2},
Mark Montine\textsuperscript{4},
Derek H. Oakley\textsuperscript{2},
Christine L. Mac Donald\textsuperscript{3},
C. Dirk Keene\textsuperscript{4},
Bradley T. Hyman\textsuperscript{2},
Bruce Fischl\textsuperscript{1},
Juan Eugenio Iglesias\textsuperscript{1,5,6}
\\
\bigskip
\textbf{1} Martinos Center for Biomedical Imaging, Massachusetts General Hospital and Harvard Medical School, Boston, Massachusetts, United States
\\
\textbf{2} Massachusetts Alzheimer's
Disease Research Center, Massachusetts General Hospital and Harvard Medical School, Boston, Massachusetts, United States
\\
\textbf{3} Department of Neurological Surgery, University of Washington School of Medicine, Seattle, WA
\\
\textbf{4} Department of Laboratory Medicine and Pathology, University of Washington School of Medicine, Seattle, Washington, United States
\\
\textbf{5} Computer Science and Artificial Intelligence Laboratory, Massachusetts Institute of Technology, Boston, Massachusetts, United States
\\
\textbf{6} Hawkes Institute, University College London, London, United Kingdom

\bigskip
* jwilliamsramirez[at]mgh.harvard.edu
\end{flushleft}
\section*{Abstract}
Advances in image registration and machine learning have recently enabled volumetric analysis of \emph{postmortem} brain tissue from conventional photographs of coronal slabs, which are routinely collected in brain banks and neuropathology laboratories around the world. One caveat of this methodology is the requirement of segmentation of the tissue from the background and out-of-slice tissue in photographs, which currently requires laborious manual intervention. In this article, we present a deep learning model to automate this process. The automatic segmentation tool relies on a U-Net architecture that was trained with a combination of 1,414 manually segmented images of both fixed and fresh tissue, from specimens with varying diagnoses, photographed at two different sites. Automated model predictions on a subset of photographs not seen in training were analyzed to estimate performance compared to manual labels, including both inter- and intra-rater variability. Our model achieved a median Dice score over 0.98, mean surface distance under 0.4~mm, and 95\% Hausdorff distance under 1.60~mm, which approaches inter-/intra-rater levels. Our tool is publicly available at \url{surfer.nmr.mgh.harvard.edu/fswiki/PhotoTools}.

\section*{Introduction}
\subsection*{Motivation}

The fundamental mechanisms underlying many neurodegenerative diseases remain poorly understood, largely due to their molecular complexity, clinical comorbidities, overlapping phenotypes, and pathological heterogeneity~\cite{jellinger2010basic,armstrong2005overlap}. As the discipline dedicated to studying and diagnosing diseases of the nervous system through the examination of brain tissue, Neuropathology plays a central role in addressing these challenges.

While certain aspects of neuropathology directly support clinical care, such as guiding therapeutic decisions through diagnosis and prognosis (e.g., tissue biopsies)\cite{ahn2010endoscopic,vital2014clinical,weis2011processing}, much of the research aimed at understanding disease pathogenesis, progression, and characterization is conducted \textit{postmortem} via autopsy\cite{sejda2022central,oura2024forensic}. Brain autopsy is particularly critical due to the limited accessibility of brain tissue \textit{ante mortem} and the unique insights gained through histological analysis. Moreover, for many brain diseases, reliable \textit{in vivo} biomarkers remain lacking, making definitive diagnosis during life difficult or impossible. As a result, neuropathological confirmation at autopsy remains the gold standard for diagnosis~\cite{hilton2015neuropathology}.

During autopsy, the brain is carefully extracted from the cadaver for subsequent analysis by dissection. Depending on the intended downstream applications, the tissue may be either frozen or chemically fixed (e.g., in formalin)\cite{adams1982atlas1}. Dissection typically begins with a gross examination and photographic documentation of the specimen following extraction and fixation, followed by sectioning into slabs. The cerebrum, cerebellum, and brainstem are slabbed separately, and the attending neuropathologist records any macroscopic abnormalities suggestive of underlying pathology\cite{adams1982atlas2}. Based on the slab size and features of interest, tissue may then be further subdivided into blocks for histological or molecular processing.

Neuropathologists now have an increasing array of tools for tissue analysis, ranging from conventional histological stains~\cite{garman2011histology} to advanced molecular techniques, including RNA sequencing and multi-omics approaches such as spatial transcriptomics applied to frozen tissue~\cite{grima2022simultaneous}. There is growing interest in integrating these molecular and histopathological data with neuroimaging modalities obtained \emph{antemortem} (e.g., magnetic resonance imaging, computed tomography, or positron emission tomography) to identify imaging biomarkers that could ultimately enable diagnosis and monitoring  \emph{in vivo}~\cite{Harsha_eLife.91398, tregidgo20203d}.

Magnetic resonance imaging (MRI) is widely regarded as the gold standard for morphometric analysis of brain specimens due to its high soft tissue contrast, excellent spatial resolution, and non-invasive nature. However, \emph{antemortem} MRI is often unavailable or acquired long before death, during which time substantial anatomical changes may occur, limiting its utility for direct comparison with postmortem histopathology.
Cadaveric MRI presents a viable alternative, offering the ability to scan intact brains shortly after death. However, unless the brain has been fixed, MRI requires rapid acquisition prior to tissue degradation that could compromise downstream analyses~\cite{guo2025single}. Such a capability is typically restricted to specialized facilities equipped with dedicated research scanners and appropriate infrastructure. An additional option is \emph{ex vivo} MRI, which provides greater flexibility in scheduling but necessitates specialized expertise and customized protocols that may not be readily available~\cite{edlow20197}.
Despite these advances, most brain banks and neuropathology laboratories worldwide lack access to MRI scanning resources altogether. While positron emission tomography (PET) enables functional imaging capabilities not possible with MRI, it is restricted to \emph{in vivo} use and remains limited in availability.

Recently, we proposed a cost-effective alternative to MRI based on dissection photography~\cite{Harsha_eLife.91398}. Nearly all brain banks and neuropathology laboratories routinely capture photographs of coronal tissue slabs during dissection for documentation purposes~\cite{rampy2015pathology}. Our method leverages these standard photographs of the cerebral hemispheres to reconstruct 3D volumes, which are then segmented into 11 distinct regions of interest (ROIs) per hemisphere using a neural network.
This approach provides a practical and accessible alternative to both \emph{in situ} and \emph{ex vivo} MRI, significantly reducing cost and acquisition time. Moreover, it enables retrospective analyses of existing datasets, unlocking the potential of legacy collections for new research applications. Even in settings where MRI or PET imaging is available, gross dissection photographs can serve as a steppingstone between the macroscopic detail of \emph{in vivo} imaging and the microscopic resolution of histology~\cite{PICHAT201873}.

\begin{figure}[t!]
    \centering
    \includegraphics[width=\textwidth]{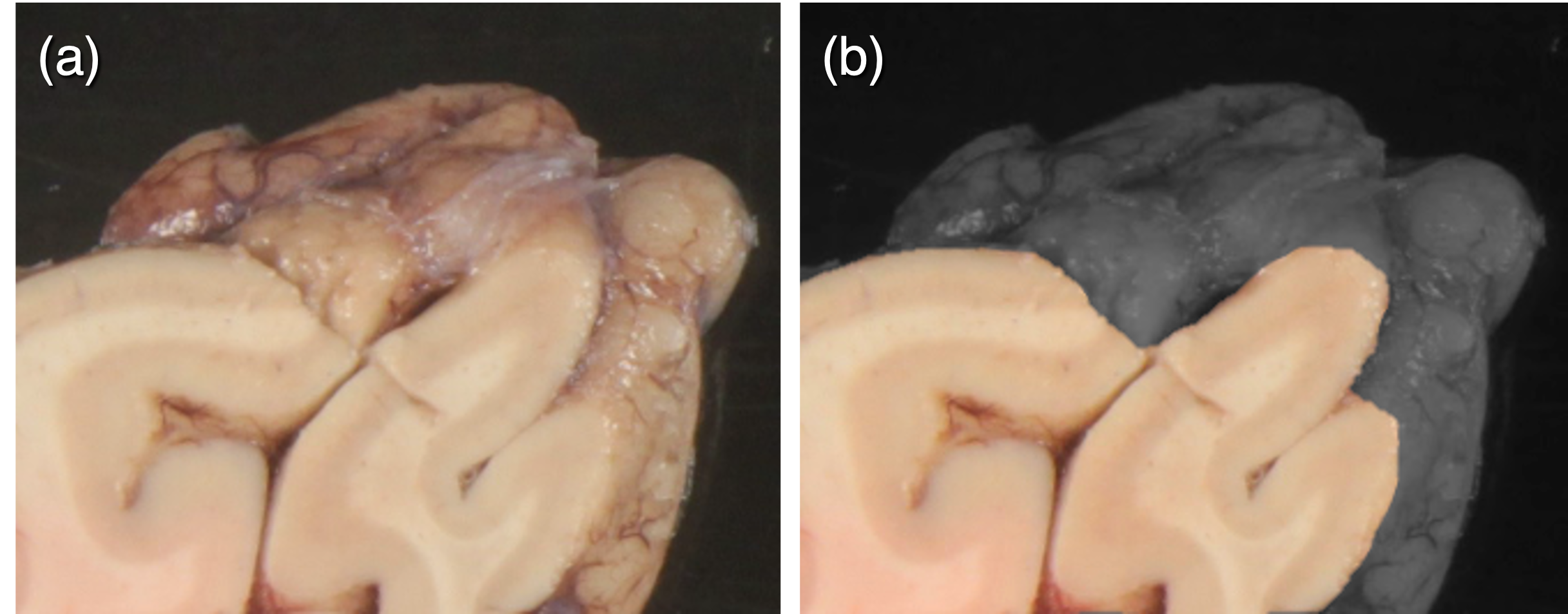}
    \caption{Close-up of a dissection photograph (a) and its corresponding reference segmentation (b). The segmentation excludes the cortical surface (shown in gray), which lies outside the plane of the block face and is not relevant for downstream analysis. Segmenting this region is often time-consuming, as simple thresholding (effective for other parts of the image) fails to distinguish it accurately. Therefore, manual tracing is typically required, making the process labor-intensive. Consequently, fully automated segmentation methods are highly desirable to improve efficiency and scalability.}
    \label{fig:photo-mask}
\end{figure}

A major challenge in scaling 3D analysis of dissection photography is the need for segmenting brain tissue from the background and out-of-slice tissue, which currently requires labor-intensive manual annotation. This is primarily due to the presence of the cortical surface in the images (Figure~\ref{fig:photo-mask}), which is not part of the slab face and should be excluded from downstream analyses~\cite{Harsha_eLife.91398}. Segmenting these non-relevant regions is particularly time-consuming, as conventional thresholding techniques (most often effective against high-contrast backgrounds) fail to accurately distinguish the cortical rim from relevant tissue. As a result, manual delineation is often necessary.
Automating this segmentation step would significantly reduce the manual workload and enhance the scalability and reproducibility of the pipeline, making high-throughput analysis of dissection photographs more feasible.

\subsection*{Contribution}
In this work, we leverage modern deep learning techniques to develop and publicly release an automated segmentation tool for coronal slab photographs. Our method is based on the nnU-Net framework~\cite{isensee2021nnu}, which combines heuristic-based configuration and internal validation to automatically adapt a U-Net architecture~\cite{ronneberger2015u} to a given segmentation task. nnU-Net has demonstrated state-of-the-art performance across a wide range of 2D and 3D medical imaging challenges~\cite{isensee2021nnu,isensee2024nnu}.

We trained our model using  1,414 real photographs from three sources, comprising fresh and fixed tissue. Our experiments show that the model achieves segmentation performance on par with manual annotations. Furthermore, even when manual corrections are necessary, automated predictions substantially reduce the time required by expert labelers, improving overall labeling efficiency. Importantly, the segmentation tool is publicly available and has been integrated into the FreeSurfer software suite~\cite{fischl2012freesurfer}, facilitating broad adoption within the neuroimaging and neuropathology communities.

\subsection*{Further related work}
Automated segmentation of ROIs from medical images is a vast field, as segmentation is a prerequisite for important downstream analyses (e.g., volumetry or shape analysis), and manual segmentation requires specific expertise, is not reproducible~\cite{pham2000current,liu2021review}, and can be highly time consuming -- particularly when performed at scale. In the context of neuropathology, classical methods have traditionally relied on handcrafted features and algorithmic approaches to segment tissue from background, delineate cellular boundaries, and classify cell types. Techniques such as thresholding, edge detection, region growing, and morphological operations have been widely used due to their simplicity and interpretability~\cite{gonzalez2009digital}. More advanced methods, including watershed algorithms~\cite{vincent1991watersheds} and active contours (snakes~\cite{kass1988snakes}), have enabled finer delineation of complex anatomical features, especially in histological and microscopy images. These approaches often leveraged highly-specific domain knowledge (e.g., staining characteristics, spatial organization, etc.) to enhance segmentation accuracy~\cite{meijering2012cell}. While effective in controlled settings, classical methods typically struggle with variability in staining, tissue artifacts, and complex cellular morphology, motivating the transition towards data-driven techniques based on machine learning for more robust and accurate automated analysis.

Deep learning has rapidly advanced computational neuropathology. Convolutional neural networks (CNNs~\cite{lecun1989backpropagation,lecun2002gradient}) and more recently transformers~\cite{vaswani2017attention}
enabled accurate and scalable solutions to tasks such as tissue segmentation, cell detection, or cell-type classification. In segmentation, architectures like the ubiquitous U-Net~\cite{ronneberger2015u} enable automatic delineation of tissue compartments or nuclei with high accuracy. Detection models, including region-based CNNs, focus on identifying individual cells or pathological structures, while classification models typically operate on image patches or single-cell representations to assign cell types or disease labels~\cite{gamper2019pannuke}. More recently, foundation models have introduced new levels of generalizability. Promptable segmentation models such as SAM~\cite{kirillov2023segment}, MedSAM~\cite{ma2024segment}, and ScribblePrompt~\cite{wong2024scribbleprompt}, can adapt to a wide range of biomedical images with minimal task-specific training. In digital pathology, large-scale pretrained models have recently demonstrated strong generalizability and performance across tasks~\cite{xu2024whole}.

In niche applications like human brain slab photography, dedicated segmentation methods remain scarce. However, existing machine learning-based segmentation approaches are highly effective and can be adapted to this specific problem, provided sufficient annotated training data are available. Despite the growing popularity of vision transformers in medical imaging \cite{han2022survey,khan2022transformers}, recent evidence demonstrates that well-designed U-Net architectures can still match or even surpass the performance of transformer-based models \cite{isensee2024nnu}. Designing an optimal U-Net model involves numerous decisions regarding architecture and hyperparameters, which can be challenging to generalize across different datasets and tasks. The widely adopted nnU-Net framework \cite{isensee2021nnu} addresses this challenge by combining heuristic rules and internal validation to automate these design choices. Heuristics, based on dataset properties such as image resolution and dimensionality, guide most configuration parameters and have been shown to generalize effectively across diverse biomedical imaging tasks. Internal validation is reserved primarily for selecting the best ensembling and post-processing strategies, as these choices are less predictable from dataset characteristics alone.


\section*{Material and methods}
\subsection*{Datasets}

Three photographic datasets were obtained from two distinct sites, each using different imaging setups and tissue types (e.g., single hemisphere vs. whole brain, frozen vs. fixed). All specimens were sectioned in the coronal plane in an anterior-to-posterior orientation and subsequently photographed using an overhead camera. Each image contains a variable number of coronal slabs, imaged from the posterior side, and includes a spatial reference object (e.g., rulers, grids, or fiducial markers) to enable pixel size scaling and, when possible, perspective correction.

\subsubsection*{Massachusetts Alzheimer's Disease Research Center (MADRC) (733 total photographs)}
This dataset includes photographs of fixed tissue slabs from 22 whole brains, 66 left hemispheres, and 41 right hemispheres (65 males, 64 females), collected at the MADRC, affiliated with Massachusetts General Hospital. The cohort primarily consists of individuals diagnosed with neurodegenerative diseases and is skewed toward older donors (average age at death: 74$\pm$14~years). Over the data collection period, several slabbing protocols were employed, resulting in variability such as differences in slab thickness. Following dissection, slabs were photographed with a ruler placed in the frame to provide a reference for pixel size estimation\cite{gazula2023machine}. An example image from this dataset is shown in Figure~\ref{fig:example_setups}a.

\subsubsection*{University of Washington, Department of Laboratory Medicine and Pathology }
Two additional datasets (one using fresh tissue and one using fixed tissue) were acquired from the Department of Laboratory Medicine and Pathology at the University of Washington (UW). For both datasets, coronal slabbing was performed using a modified meat slicer that produced consistent 4~mm slices. Prior to slicing, specimens were embedded in dental alginate to stabilize the tissue and minimize deformation during cutting.

\begin{figure}[t!]
    \centering
    \includegraphics[width=\textwidth]{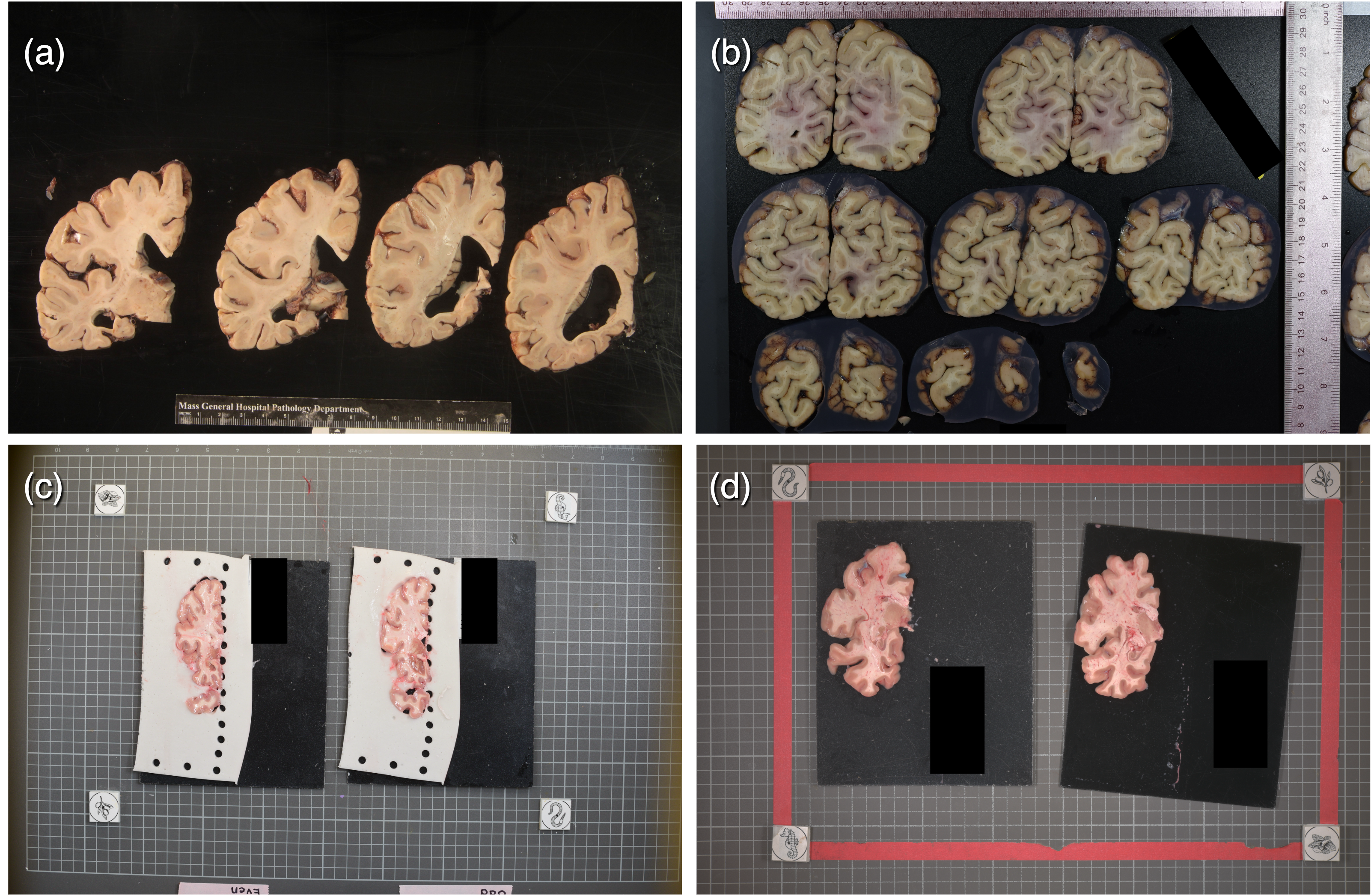}
    \caption{Example dissection photography setups across the three photographic datasets used in this study.
(a)~Photograph from the MADRC dataset showing formalin-fixed slabs from a whole brain, with a ruler for scale calibration.
(b)~Image from the UW-fixed dataset, also depicting formalin-fixed whole-brain slabs with two orthogonal rulers included for spatial reference.
(c,d)~Images from the UW-fresh dataset showing coronal slabs from single hemispheres photographed on two slightly different tables. Four fiducial markers placed in a rectangular configuration are used for pixel size and perspective correction.}
    \label{fig:example_setups}
\end{figure}

\begin{itemize}
\item\textbf{UW-fixed (218 total photographs):}
This dataset consists of dissection photographs from 28 whole brains (17 males, 11 females), fixed in 10\% neutral buffered formalin. Similar to the MADRC dataset, many donors in this cohort had neurodegenerative conditions such as Alzheimer's disease, Lewy body dementia, or Parkinson’s disease (average age at death: 67$\pm$23~years). Each image includes two orthogonal rulers to enable accurate pixel size calibration. A representative image is shown in Figure~\ref{fig:example_setups}b.
\item\textbf{UW-fresh (681 total photographs):}
This dataset comprises coronal slabs from 38 single hemispheres (19 from males, 19 from females), with donors having an average age at death of 48$\pm$17~years.  This datasets enables us to test our algorithm under more challenging conditions, since fresh tissue (which allows for analyses that are not possible with fixed tissue due to damage to proteins and genetic material, such as  RNA sequencing and transcriptomics) suffers from much stronger nonlinear deformation than fixed specimens. Four fiducial markers (on the corners of a rectangle of known dimensions) were used as references for perspective and pixel size estimation~\cite{Harsha_eLife.91398}. Two slightly different tables (shown in Figures~\ref{fig:example_setups}c and~\ref{fig:example_setups}d) were used to acquired this dataset.
\end{itemize}

\subsection*{Methods}
\subsubsection*{Image preprocessing}
Calibrating pixel size in medical image segmentation is essential to ensure that spatial measurements reflect true anatomical dimensions. Eliminating pixel size variability enables better generalization across photographs from different sites, resolutions, or zoom levels. Our preprocessing followed \cite{Harsha_eLife.91398} to resample all images to the target resolution (in practice, 0.1~mm). In the MADRC and UW-fixed datasets, we manually clicked on two points along reference rulers and specified the distance between them in physical units (mm). This enabled us to estimate an isotropic scaling factor that was used for resampling. In the UW-fresh dataset, we automatically detected the fiducials with a combination of SIFT~\cite{lowe1999object} and RANSAC~\cite{fischler1981random} algorithms, and used the coordinates to estimate a homography that was used to resample perspective-corrected images at the target resolution.

\subsubsection*{Manual delineation}
Each photograph was manually annotated once by one of four labelers (JWR, LDB, RH, EGP) using open-source photo editing software (GIMP v2.8). Annotators removed all tissue outside the slab surface, including the cortical surface (Figure~\ref{fig:photo-mask}), which is not part of the slab face and should not be considered in subsequent analyses. Therefore, precise masking of the cortical surface in the training data is important to train a model that can accurately segment it at test time -- particularly in datasets with thick slabs that expose larger portions of the cortex, as in Figure~\ref{fig:photo-mask}a. Manual segmentation combined thresholding (which efficiently captures the majority of boundaries) with subsequent manual boundary tracing to refine the contours where needed, typically along the cortical surfaces described earlier.

\subsubsection*{Model Training and Architecture}
We trained our neural network using 733 photographs from the MADRC dataset and 681 from the UW-fresh dataset. The UW-fixed dataset was reserved exclusively for testing, allowing us to evaluate the model’s ability to generalize to previously unseen data. We note that nnU-net internally partitions the training data into training and validation splits, to combat overfitting. 

To optimize inference speed while preserving anatomical detail, all images were downsampled to a resolution of 0.5mm/pixel, striking a balance between spatial resolution and computational efficiency. The average image dimensions used during training were 440×~673 pixels.

The final U-Net architecture (automatically configured by the nnU-Net framework) consisted of seven stages with 32, 64, 128, 256, 512, 512, and 512 feature channels, respectively. Model training was performed over five cross-validation folds, each trained for 1,000 epochs, and required a total of 58.3 hours on an NVIDIA RTX A6000 GPU.

\subsubsection*{Evaluation metrics}
To evaluate automated segmentations with respect to the reference masks, we use modified versions of three standard metrics: Dice overlap, average symmetric surface distance, and the 95\% Hausdorff Distance (HD95). The modification lies in the fact that we only compute the metric over pixels within 10~mm of the (morphologically filled) reference segmentations. This is to reflect the fact that we do not mind false positives far away from the target slabs (e.g., due to the presence of non-target slabs, as in Figure~\ref{fig:examples}f or Figures \ref{fig:s3} and \ref{fig:s4} in the supplement), and which can be easily masked during postprocessing. 

\subsection*{Ethics Statement}
Human tissue used in this study was obtained in accordance with institutional and national ethical guidelines. Procedures for tissue collection and processing at Massachusetts General Hospital were reviewed and approved by the Partners Human Research Committee institutional review board under protocol 1999P009556. All \textit{postmortem} specimens were obtained with informed consent from next of kin. At the University of Washington, tissue processing and collection involves the receipt and analysis of de-identified information and specimens from deceased individuals which were acquired from a biorepository.  Consent for the use of information and specimens from the deceased individuals was obtained either from the individual while they are alive, or from the individual's legally authorized representative. Procurement and banking practices of the biorepository are informed by the US Revised Uniform Anatomical Gift ACT 2006 (Last Revised or Amended in 2009) and Washington Statute Chapter 68.64 RCW. We work closely with the UW School of Medicine Compliance Office on our consent forms and HIPAA compliance. All materials are collected under informed consent.


\section*{Experiments and results}

\subsection*{Manual segmentation: time efficiency and inter-/intra-rater variability}

To contextualize the performance of the automated method, we first conducted an experiment involving only human labelers, with two main objectives: to assess labeling time efficiency (i.e., the time required for manual annotation), and to evaluate variability both within and between annotators (i.e., intra- and inter-rater variability).

Labeling time was measured on a subset of 20 photographs. On average, manual annotation took 3 minutes and 29 seconds per tissue slab, implying that a typical photograph containing three slabs required approximately 10 minutes to label. Extrapolating this estimate to the entire training dataset of 1,414 images yields a total of 214.9 hours of continuous manual labeling. These results underscore the potential of the automated approach to substantially reduce expert annotation time and associated resource costs.

To assess annotation variability, 20 photographs (10 from the MADRC dataset and 10 from the UW-fresh dataset) were randomly selected and annotated twice by two labelers (JWR and LDB), with a one-week interval between sessions to minimize memory bias. Figure~\ref{fig:All_metrics} presents the Dice scores, average symmetric surface distances, and HD95 for both intra- and inter-rater comparisons. For inter-rater evaluation, results were averaged across all four possible labeler pairings. All metrics indicated excellent agreement, with median Dice scores exceeding 0.98, average surface distances below 0.5~mm, and HD95 values under 2~mm. Notably, the inter-rater variability was comparable to the intra-rater variability of the more variable annotator, demonstrating strong overall consistency and reliability in the manual annotations.

\subsection*{Automated segmentation: in- and out-of-distribution images}

We computed Dice scores and surface distance metrics for two evaluation settings: \textit{(i)}~89 photographs from the MADRC dataset and 70 from the UW-fresh dataset that were withheld during training, and \textit{(ii)}~the full UW-fixed dataset, comprising 218 photographs. The first setting evaluates model performance on in-distribution data, while the second assesses generalization to out-of-distribution images.

\begin{figure}[t!]
    \centering
    \includegraphics[width=\textwidth]{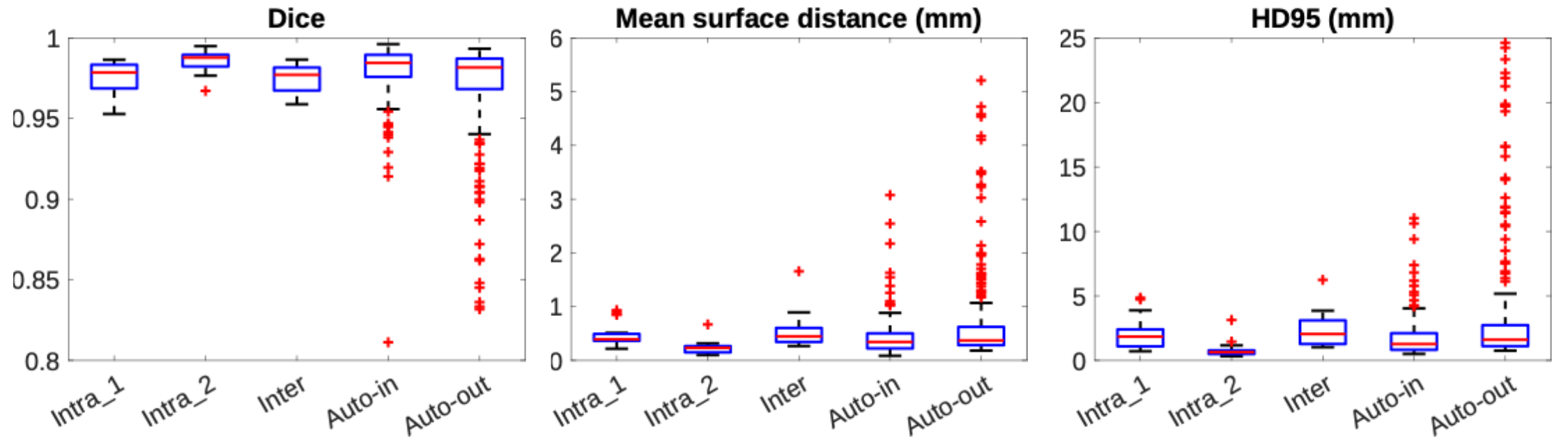}
    \caption{Box plots for Dice scores, mean surface distance, and HD95 for manual and automated methods. ``Intra-1'': intra-rater variability of Labeler~1, measured on 20 images. ``Intra-2'': intra-rater variability of Labeler~2 on the same 20 images. ``Inter'': inter-rater variability of the same 20 images. ``Auto-in'': accuracy of the proposed automated method on the in-distribution data (159 photographs).  ``Auto-out'': accuracy of the proposed automated method  on the out-of-distribution data (218 photographs).}
    \label{fig:All_metrics}
\end{figure}

Some segmentations are shown in Figure~\ref{fig:examples} and in the supplement Figure~\ref{fig:s1} to~\ref{fig:s6}, whereas quantitative results are shown in Figure~\ref{fig:All_metrics}. On in-distribution data, the model achieved excellent performance, with a median Dice score of 0.985, a mean surface distance of 0.34~mm, and a HD95 of 1.27~mm. These values closely approach the intra-rater variability of the more consistent human annotator (noting that this comparison involves different image samples). Using a very conservative threshold of 2~mm for the mean surface distance to define outliers, the in-distribution outlier rate was just 2\%.

On the out-of-distribution UW-fixed dataset, the model maintained strong performance, with a median Dice score of 0.981, a mean surface distance of 0.37~mm, and an HD95 of 1.60~mm, closely matching results on in-distribution data. These results indicate that the model generalizes well to previously unseen data in the majority of cases.
As expected with domain shifts, variability increased slightly: approximately 6\% of images exhibited errors greater than 2~mm, as seen in the wider spread of the box plots across all metrics. However, this increase is modest in absolute terms, particularly given the relatively stringent threshold used to define outliers. Overall, the model demonstrates robust generalization with only minimal degradation in performance on out-of-distribution samples.
    
\begin{figure}[t!]
    \centering
    \includegraphics[width=\textwidth]{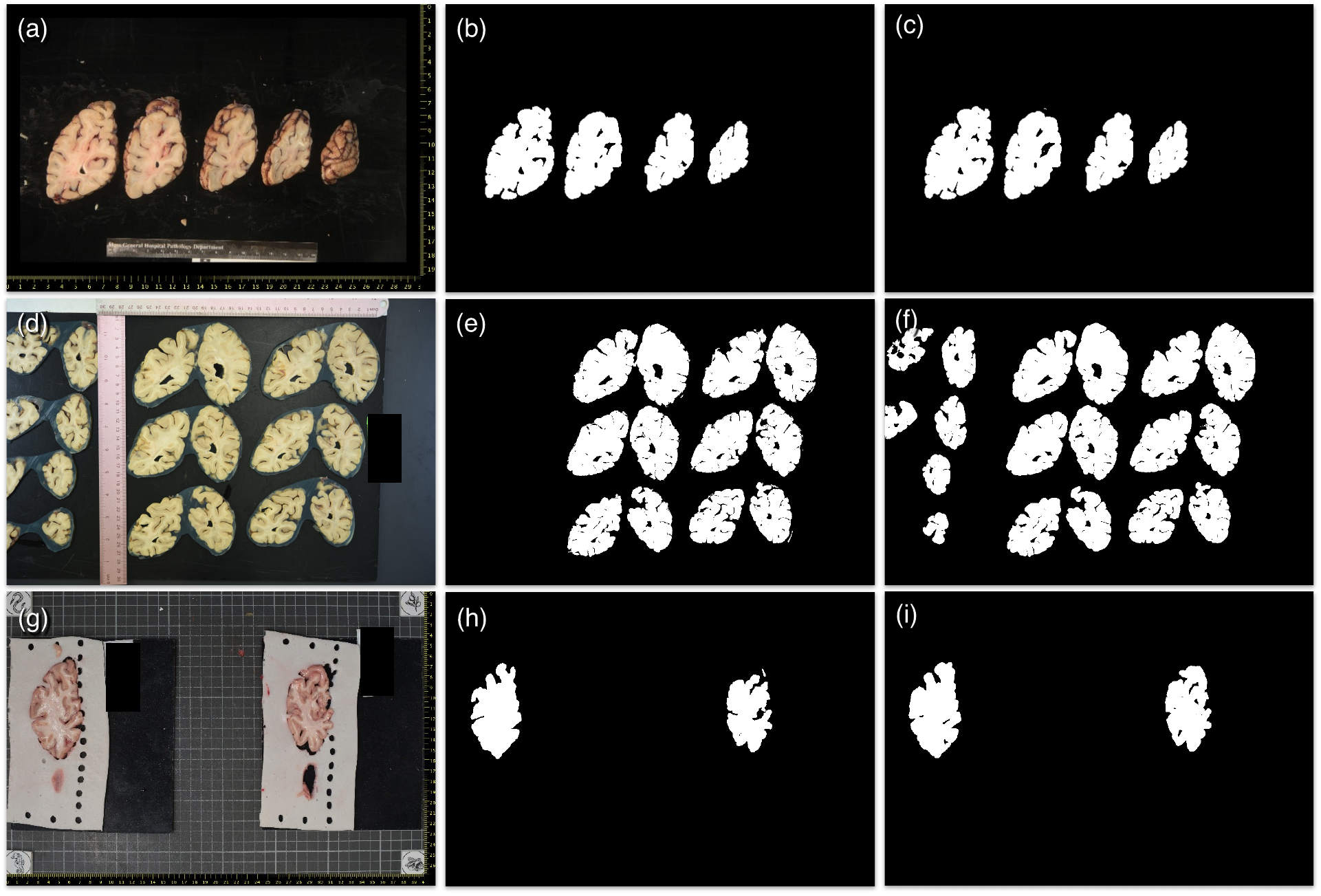}
    \caption{(a)~Sample photograph from the MADRC dataset. (b)~Corresponding reference mask. (c)~Automated segmentation. (d-f)~Example from UW-fixed dataset; note the non-target slabs leading to irrelevant false positives that are factored out of our accuracy metrics. (g-h)~Example from the UW-fresh dataset. A more comprehensive set of examples can be found in the supplement Figure~\ref{fig:s1} to~\ref{fig:s6}}.
    \label{fig:examples}
\end{figure}

\section*{Discussion}
This work introduces a fully automated segmentation tool for brain dissection photographs, offering a practical and scalable alternative to resource-intensive annotation workflows. By leveraging the nnU-Net framework, our method achieves a level of accuracy that is comparable to that of expert human annotators. It yields strong performance on in-distribution data, and generalizes effectively to out-of-distribution images from a previously unseen site.

Despite these encouraging results, a key limitation of this study is the diversity of the data sources. All training and evaluation were conducted on photographs acquired from only three sites. While the diversity in dissection protocols, imaging setups, and tissue conditions (e.g., fixed vs. fresh) helps build robustness, it is still limited in scope. Further validation on additional datasets (particularly from institutions with different photographic and tissue-handling practices) will be important to assess and potentially improve the generalizability of the tool.

To this end, future work may explore additional strategies to enhance robustness across domains. For example, synthetic data generation with domain randomization could further expose the model to a broader range of appearances and imaging conditions. Incorporating style transfer or domain adaptation techniques may also reduce sensitivity to site-specific artifacts. Another potential direction is to use lightweight image-level quality control (QC) mechanisms, such as binary pass/fail ratings, to complement traditional spatial metrics such as Dice and surface distance. Such QC frameworks may provide more intuitive assessments of segmentation reliability in large-scale, real-world workflows.

Ultimately, the practical impact of this tool lies in its potential to dramatically reduce the time burden of manual annotation. As we estimated, segmenting a typical dissection photograph takes approximately 10 minutes by hand. Extrapolated to our training dataset alone, this represents over 200 hours of expert time. Our tool can almost entirely remove this workload for most images and still substantially accelerate expert review when manual refinement is needed. This represents a significant step toward scalable, high-throughput analysis of neuropathology specimens, unlocking the potential of vast image archives that would otherwise be prohibitively time-consuming to label manually.

\section*{Acknowledgments}
This research was primarily funded by the National Institute of Aging (1R01AG070988). Additional support was provided by the National Institute of Mental Health (1RF1MH123195, 1UM1MH130981), the National Institute of Aging (1RF1AG080371), the National Institute of Biomedical Imaging and Bioengineering (1R01EB031114), and the National Institute of Neurological Disorders and Stroke (1R21NS138995). The UW BioRepository and Integrated Neuropathology (BRaIN) laboratory is supported by the National Institutes of Health (NIH) through the UW Alzheimer’s Disease Research Center (P30 AG066509), the Kaiser Permanente Washington Adult Changes in Thought (ACT) study (U19 AG066567), the BRAIN Initiative Cell Atlas Network (UM1MH134812), the Seattle Alzheimer’s Disease Brain Cell Atlas (U19AG060909), multiple collaborative projects supporting imaging, digital pathology, neuropathology, and related research tools (U24AG072458; U24NS133949; U24NS133945; U24NS135651; U01NS137500; and U01NS137484), the US Department of Defense (DoD W81XWH-21-S-TBIPH2), the Allen Institute for Brain Science, and the Nancy and Buster Alvord Endowment (to C.D.K.).
Support for this research was provided in part by the BRAIN Initiative Cell Atlas Network (BICAN) grants U01MH117023, UM1MH134812 and UM1MH130981, the Brain Initiative Brain Connects consortium (U01NS132181, 1UM1NS132358-01), the National Institute for Biomedical Imaging and Bioengineering (1R01EB023281, R21EB018907, R01EB019956, P41EB030006), the National Institute on Aging (R21AG082082, 1R01AG064027, R01AG016495, 1R01AG070988), the National Institute of Mental Health (UM1MH130981, R01 MH123195, R01 MH121885, 1RF1MH123195), the National Institute for Neurological Disorders and Stroke , (1U24NS135561-01, R01NS070963, 2R01NS083534, R01NS105820, R25NS125599), and was made possible by the resources provided by Shared Instrumentation Grants 1S10RR023401, 1S10RR019307, and 1S10RR023043. Additional support was provided by the NIH Blueprint for Neuroscience Research (5U01-MH093765), part of the multi institutional Human Connectome Project. Much of the computation resources required for this  research was performed on computational hardware generously provided by the Massachusetts Life Sciences Center (https:/ www.masslifesciences.com).
The authors would like to thank the research participants and their families without whom this work would be impossible.  

\section*{Conflicts of Interest}
Dr Hyman owns stock in Novartis; he serves on the SAB of Dewpoint and has an option for stock.  He serves on a scientific advisory board or is a consultant for AbbVie, Alexion,  Ambagon, Aprinoia Therapeutics, Arbor Bio, Arvinas, Avrobio, AstraZenica, Biogen, Bioinsights, BMS, Cell Signaling, Cure Alz Fund, CurieBio, Dewpoint, Etiome, Latus, Merck,  Novartis, Paragon, Pfizer, Sanofi, Sofinnova, SV Health, Takeda, TD Cowen, Vigil, Violet, Voyager, WaveBreak.  His laboratory is supported by research grants from the National Institutes of Health, Cure Alzheimer’s Fund, Tau Consortium, and the JPB Foundation – and sponsored research agreement from Abbvie and Sanofi. He has a collaborative project with Biogen and Neurimmune.
Dr Fischl is an advisor to DeepHealth, a company whose medical pursuits focus on medical imaging and measurement technologies. BF's interests were reviewed and are managed by Massachusetts General Hospital and Mass. General Brigham in accordance with their conflict of interest policies.

\nolinenumbers

\appendix
\renewcommand{\thefigure}{S\arabic{figure}}
\setcounter{figure}{0}
\clearpage
\section*{Supplementary Material}

\begin{figure}[htbp]
    \centering
    \includegraphics[width=\textwidth]{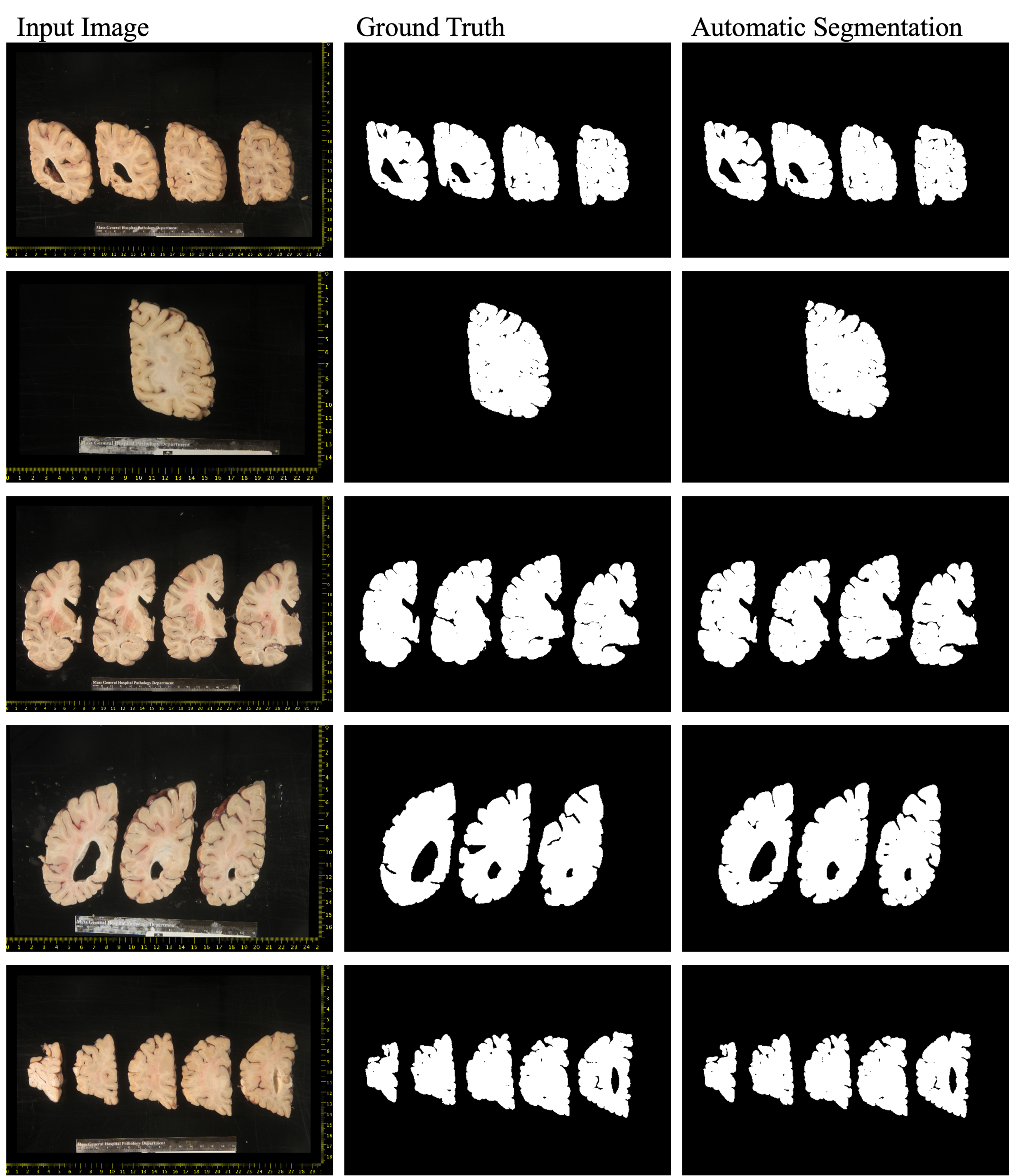}
    \caption{Additional automated segmentations for sample images from MADRC dataset (in distribution).}
    \label{fig:s1}
\end{figure}

\begin{figure}[htbp]
    \centering
    \includegraphics[width=\textwidth]{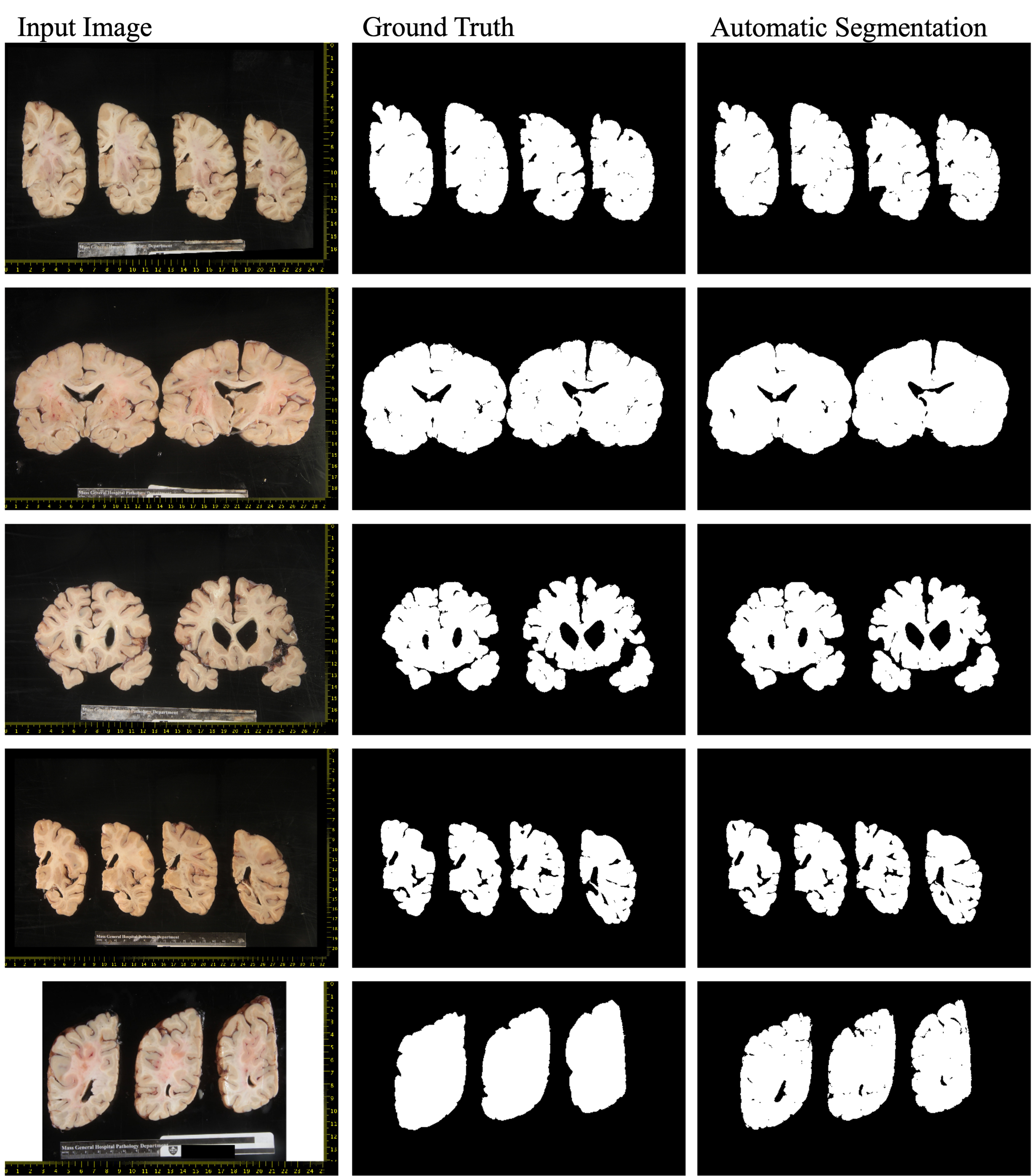}
    \caption{Additional automated segmentations for sample images from MADRC dataset (in distribution).}
    \label{fig:s2}
\end{figure}

\begin{figure}[!t]
    \centering
    \includegraphics[width=\textwidth]{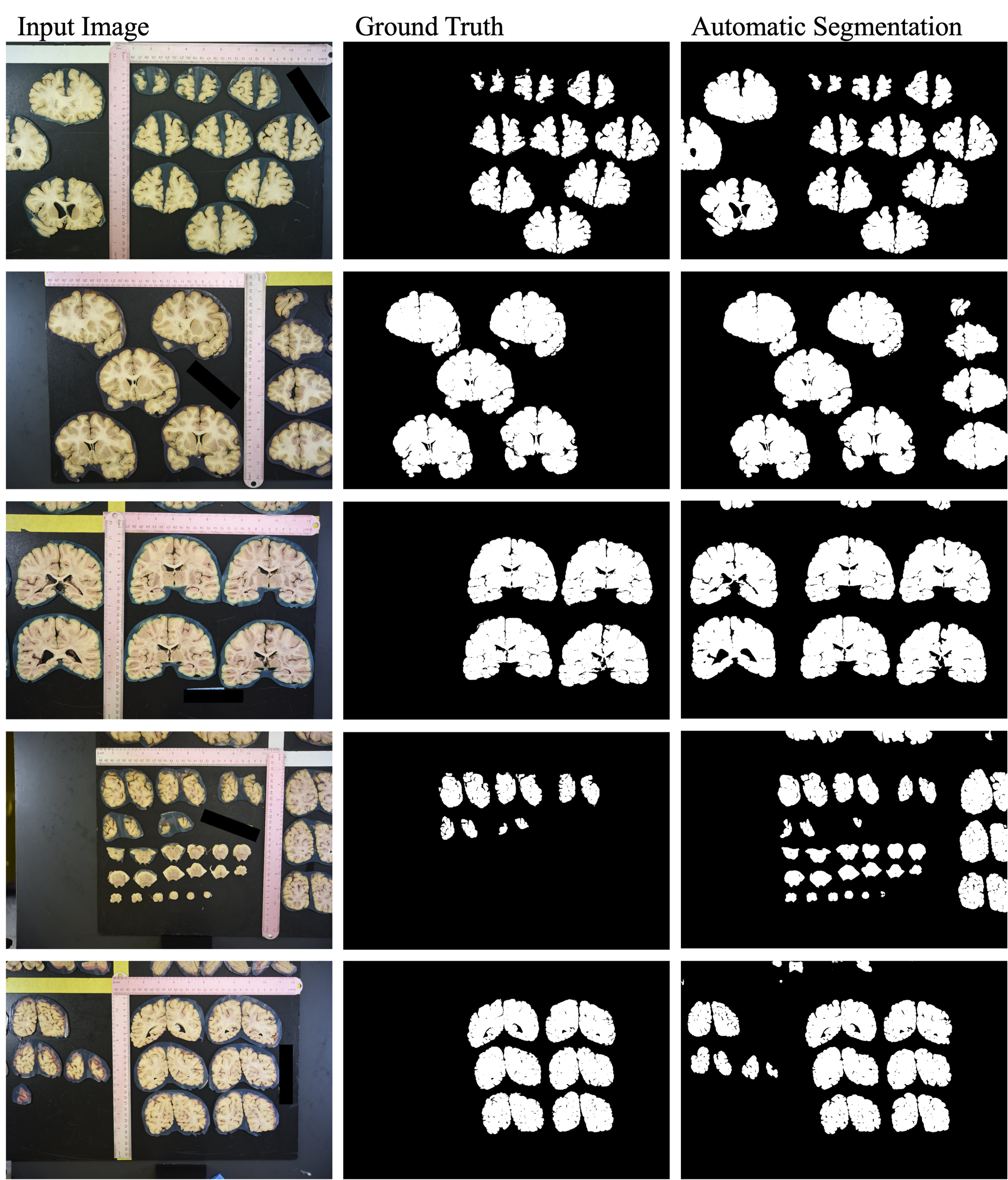}
    \caption{Additional automated segmentations for sample images from UW-fixed dataset (out of distribution). Please note that we do not mind false positives on non-target slabs, which can be easily masked during postprocessing.}
    \label{fig:s3}
\end{figure}

\begin{figure}[!t]
    \centering
    \includegraphics[width=\textwidth]{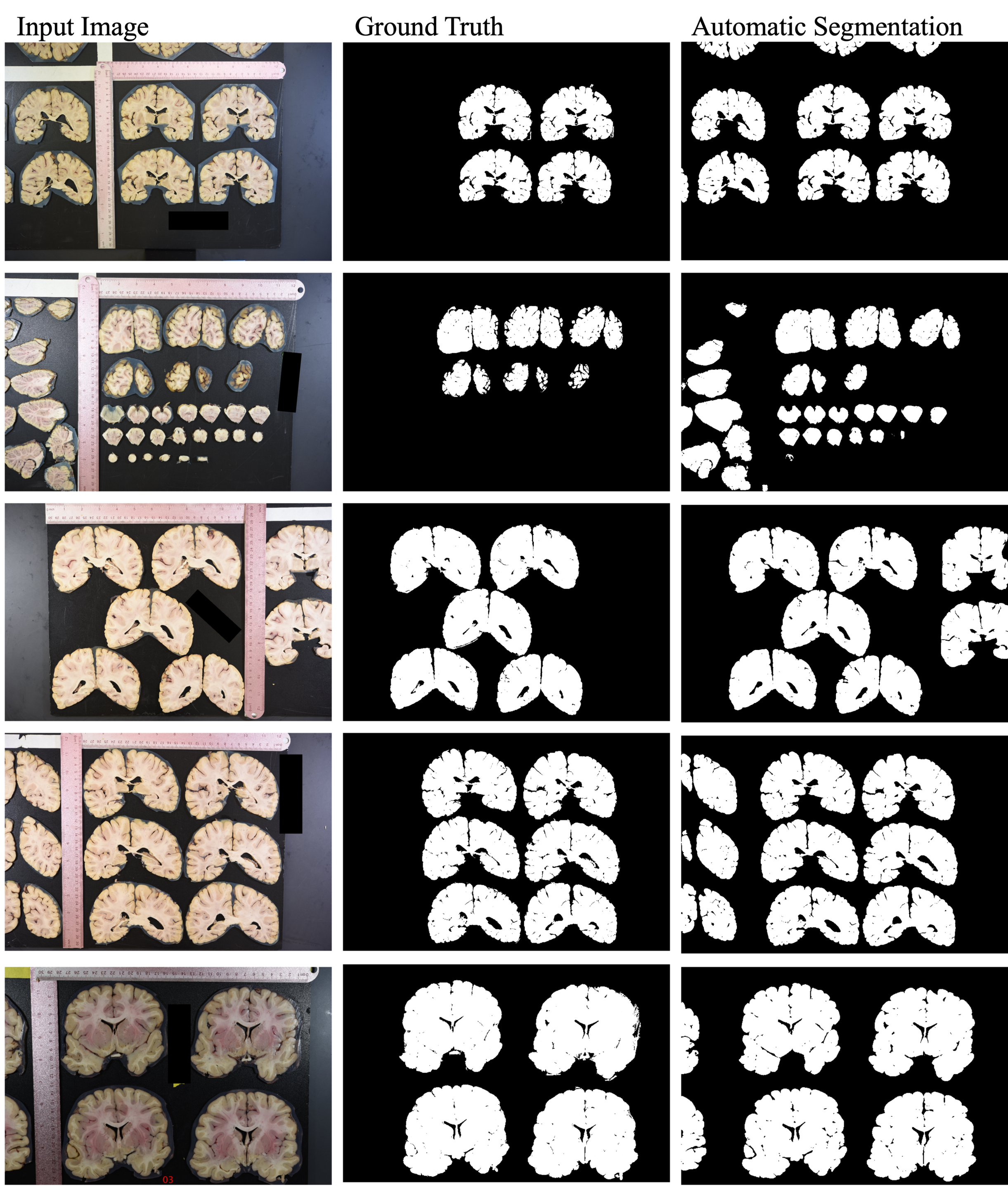}
    \caption{Additional automated segmentations for sample images from UW-fixed dataset (out of distribution). Please note that we do not mind false positives on non-target slabs, which can be easily masked during postprocessing.}
    \label{fig:s4}
\end{figure}

\begin{figure}[!t]
    \centering
    \includegraphics[width=\textwidth]{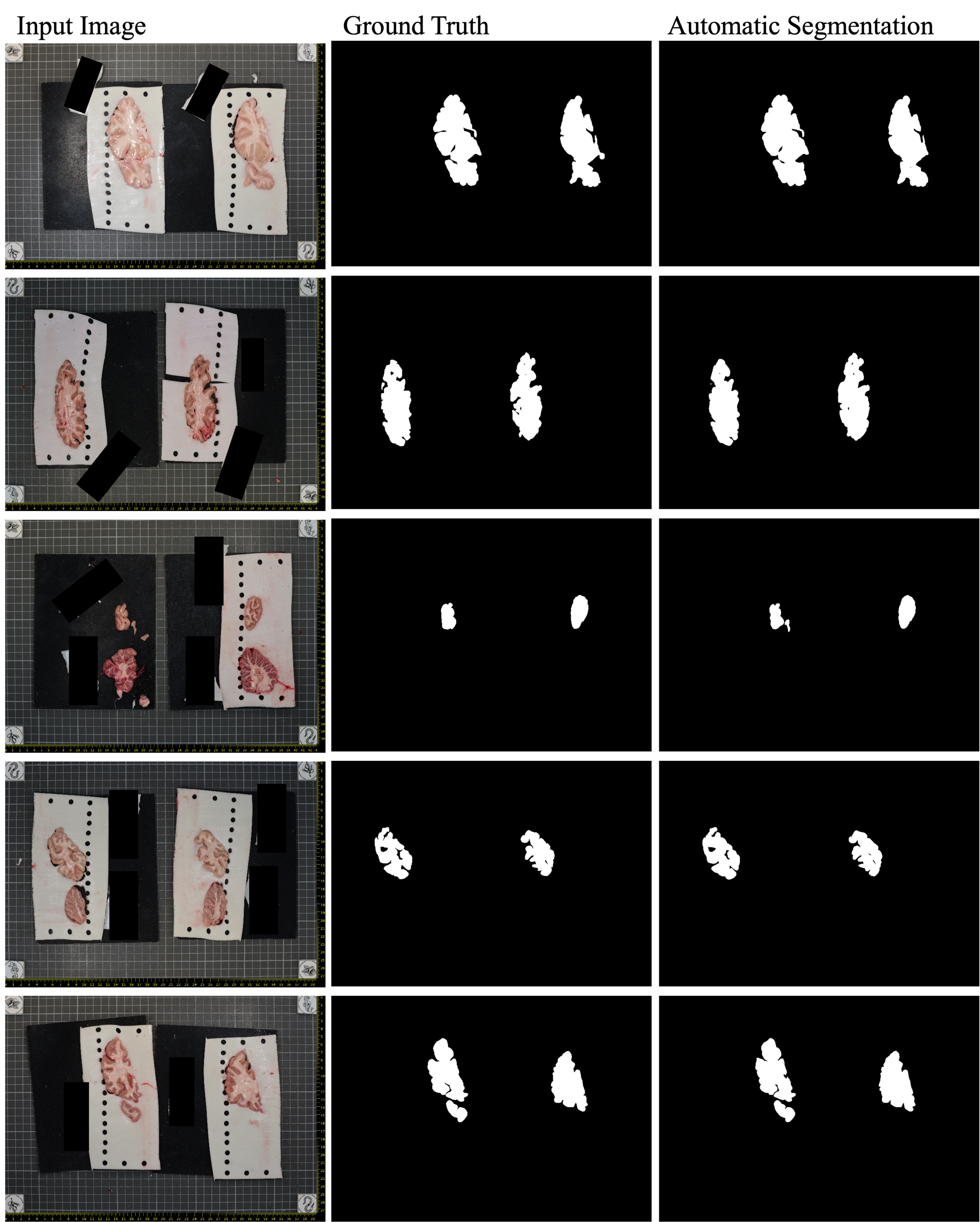}
    \caption{Additional automated segmentations for sample images from UW-fresh dataset (in distribution).}
    \label{fig:s5}
\end{figure}

\begin{figure}[!t]
    \centering
    \includegraphics[width=\textwidth]{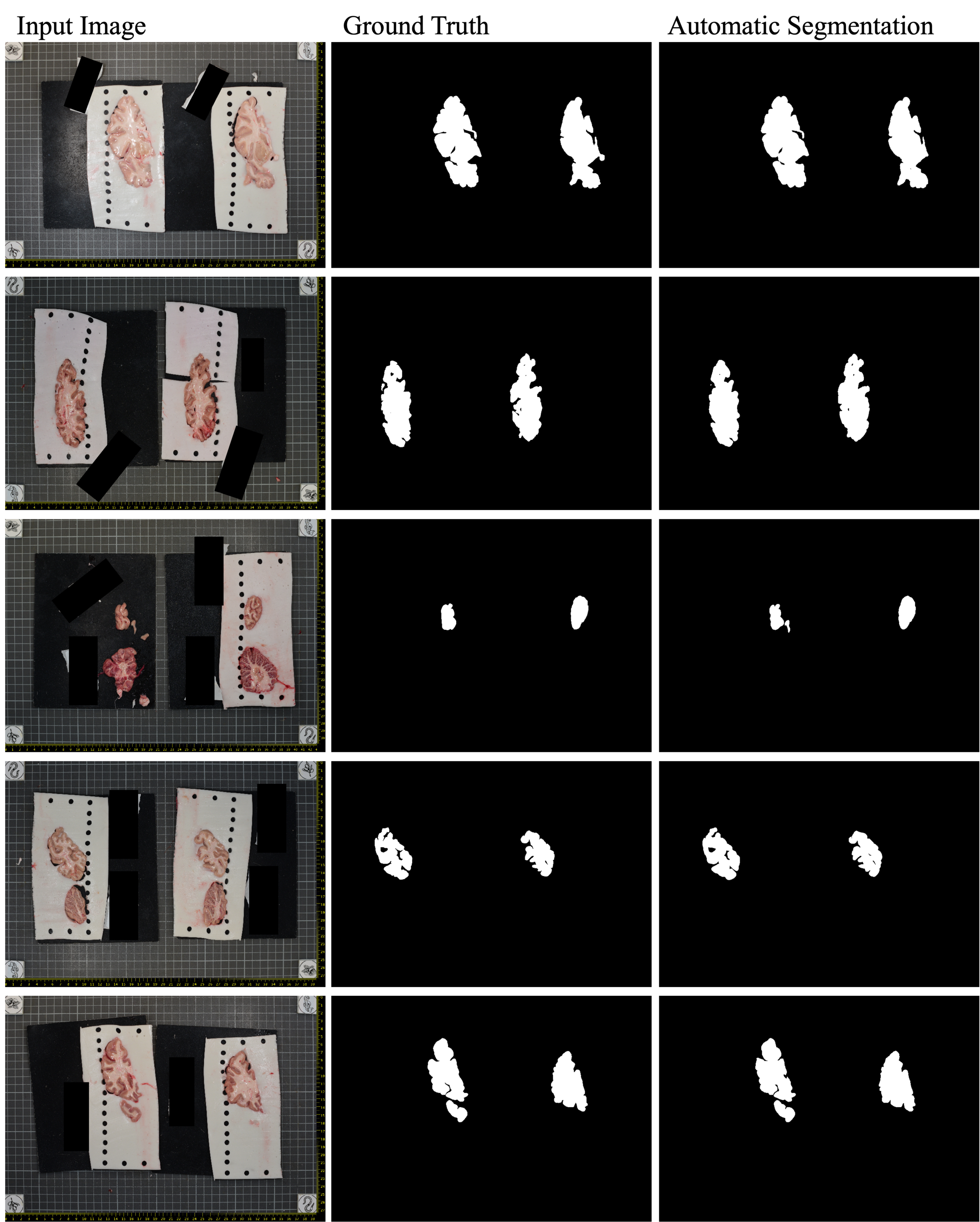}
    \caption{Additional automated segmentations for sample images from UW-fresh dataset (in distribution).}
    \label{fig:s6}
\end{figure}

\end{document}